\documentclass[conference]{IEEEtran}
\usepackage{color}
\usepackage{algorithm,algorithmicx}
\usepackage{algpseudocode,empheq,hhline}
\usepackage{amsmath,amsthm,amssymb,url,graphicx,rotating,ifthen,epsfig,array,caption,color}
\usepackage{todonotes}
\usepackage{graphicx}
\usepackage{caption}
\usepackage{subcaption}
\usepackage{balance}
\usepackage{multirow}
\usepackage{enumitem}
\def\x{\mathbf{x}}
\def\N{{\mathbb N}}
	\def\G{{\mathcal N}}
\def\P{{\cal P}}

\newcommand{\subscript}[2]{$#1 _ #2$}

\ifCLASSINFOpdf
\else
\fi
\begin{document}
%


\title{Evaluating Noisy Optimisation Algorithms:  \\ First Hitting Time is Problematic}


\def\nouse{
\author{
Simon~M.~Lucas,~\IEEEmembership{Senior Member,~IEEE,}
Jialin~Liu,~\IEEEmembership{Member,~IEEE,}
Diego~Perez-Liebana,~\IEEEmembership{Member,~IEEE}

\thanks{S. M. Lucas, J. Liu, D. Perez-Liebana are with the Department of Computer Science and Electronic Engineering, University of Essex, Colchester CO4 3SQ, UK
(e-mail: sml@essex.ac.uk; jialin.liu@essex.ac.uk; dperez@essex.ac.uk).}
}}
\author{
\IEEEauthorblockN{Simon M. Lucas}
\IEEEauthorblockA{University of Essex\\
Colchester CO4 3SQ\\
United Kingdom\\
\url{sml@essex.ac.uk}}
\and
\IEEEauthorblockN{Jialin Liu}
\IEEEauthorblockA{University of Essex\\
Colchester CO4 3SQ\\
United Kingdom\\
\url{jialin.liu@essex.ac.uk}}
\and
\IEEEauthorblockN{Diego P\'erez-Li\'ebana}
\IEEEauthorblockA{University of Essex\\
Colchester CO4 3SQ\\
United Kingdom\\
\url{dperez@essex.ac.uk}}
}
\maketitle

\begin{abstract}
A key part of any evolutionary algorithm is fitness evaluation.  
When fitness evaluations are corrupted by noise, as happens in many real-world problems as a consequence of various types of uncertainty, a strategy is needed in order to cope with this.  
Resampling is one of the most common strategies, whereby each solution is evaluated many times in order to reduce the variance of the fitness estimates.

When evaluating the performance of a noisy optimisation algorithm, a key consideration is the stopping condition for the algorithm.  A frequently used stopping condition in runtime analysis, known as ``First Hitting Time'',  is to stop the 
algorithm as soon as it encounters the optimal solution.  However, this is unrealistic for real-world problems, as if the optimal solution were already known, there would be no need to search for it.

This paper argues that the use of First Hitting Time, despite being a commonly used approach, 
is significantly flawed and overestimates the quality of many algorithms in real-world cases,
where the optimum is not known in advance and has to be genuinely searched for.
A better alternative  is to measure the quality of the solution an algorithm
returns after a fixed evaluation budget, i.e., to focus on final solution quality.  
This paper argues that focussing on final solution quality is more realistic
and demonstrates cases where the results produced by each algorithm evaluation method lead 
to very different conclusions regarding the quality of each noisy optimisation algorithm.

\end{abstract}


%
\IEEEpeerreviewmaketitle

\section{Introduction}

Noisy optimisation problems are important and commonplace in real-world applications, where evaluating the fitness of a solution may be subject to various forms of uncertainty. This paper deals with how to evaluate the performance of a noisy optimisation algorithm. The main purpose of the paper is to demonstrate that a commonly used measure, First Hitting Time (FHT), can give very misleading results: results which will not be indicative of real-world performance.  The divergence between the performance measured by FHT and the real-world utility of the algorithm may be very large, as will be shown in the results (Section \ref{sec:res}).  The problem arises when an optimisation algorithm stumbles upon the optimal solution without being confident that it has found it.  If the performance evaluator rewards an algorithm for merely visiting the optimum as opposed to returning the optimum as its best guess, then a significantly inflated estimate of the optimiser's performance may be given.


\emph{Evaluation of noisy optimisation algorithms:}
Commonly used measures of noisy optimisation algorithms are FHT when the search space is finite or countable, and the convergence rate in continuous domains.
The FHT is the first time at which an optimum\footnote{In this paper, the term ``optimum'' refers to the optimal solution, ``optimal value'' refers to the noise-free fitness value of the optimum.} is visited.  FHT is widely used to evaluate the performance of noisy optimisation algorithms in problems with discrete search space, such as various noisy versions of the OneMax problem~\cite{droste2004analysis,qian2014effectiveness,akimoto2015analysis,liu2016optimal,qian2016effectiveness}.
In the noisy case, some resampling techniques\footnote{The term ``sampling'' refers to sample the search points (solutions), while ``resampling'' stands for evaluating the identical solution more than once.} are often used aiming at cancelling the effect of noise~\cite{finck2010real,rakshit2016noisy,liu2016optimal}.
In most of the previous work~\cite{droste2004analysis,qian2014effectiveness,liu2016optimal}, the runtime analysis is based on the assumption that once the optimal solution is hit, an absorbing state is reached and the optimisation terminates.  This assumption is not realistic in real-world applications, as if the optimal solution were known, no optimisation or search would be required.

A better alternative to FHT is to measure the quality of the solution an algorithm
returns after a fixed evaluation budget, i.e., to focus on final solution quality.  
This can focus on any measure of solution quality, but in this paper
for ease of comparison and for clarity we focus on the success rate defined
as the percentage of trials that the optimiser returns the optimal solution as its final solution.

In this paper we demonstrate the impact of choosing an appropriate methodology for 
evaluating a noisy optimisation algorithm.  In particular, we do this by showing two test problems (described in Section~\ref{sec:problems}) where choosing an
inapplicable evaluation methodology can lead to the opposite conclusion compared to choosing a more applicable one.
We test two different optimisation algorithms: the Random Mutation Hill Climber (RMHC) and the (1+1)-EA under the two different evaluation methods, in order to demonstrate  that the results are not unique to a single optimisation algorithm.  These
are described in Section~\ref{sec:experiment}.
Section~\ref{sec:res} summarises the results and discusses the dramatic impact that the evaluation method can have on the estimated performance of noisy optimisation algorithms and the prediction of the optimal resampling rate.  Finally, Section~\ref{sec:conc} concludes that First Hitting Time should not be used
to evaluate the performance of a noisy optimisation algorithm.

\def\nouse{
\section{Impact of Evaluation Methodology\label{sec:impact}}

}
\section{Test Problems}\label{sec:problems}
We considered two problems: the OneMax problem corrupted by additive Gaussian noise, and an artificial game outcome optimisation problem.

\subsection{OneMax problem with additive Gaussian noise}
The $n$-bit OneMax problem corrupted by additive Gaussian noise is formalised as: 
\begin{equation}
f(\x)=\sum_{i=1}^{n} \x_i + \G(0,1), \label{eqn:NoiseFreeFitnessFunction}
\end{equation}
where $\x$ is an $n$-bit binary string, $\G(\mu,\sigma^2)$ denotes Gaussian noise with mean $\mu$ and variance $\sigma^2$. 
This problem is referred to as the \emph{Noisy OneMax} problem in the rest of the paper.

\subsection{Optimisation of game outcome}\label{sec:gp}
We also consider an artificial game outcome optimisation problem, referred to as the \emph{Noisy PMax} problem, in the rest of the paper.
The bit-string $\x$ can be:
\begin{enumerate}[label=(\subscript{P}{\arabic*})]
\item \label{itm:first} a parameter setting of an AI agent to play a given game
\item \label{itm:second} or a parameter setting of a game for a given AI agent.
\end{enumerate}
The pretext for this problem is that we are trying to optimise the winning rate for a given agent
but due to the stochastic nature of the game and / or the agent(s) playing it, the outcome of every simulation of the game is a win or a loss with a given
probability.  This leads to a very noisy optimisation problem.  

The Noisy PMax problem was directly inspired by recent work on evolving game
parameters in order to maximise the winning rate of a skilful agent over
less skilful ones \cite{Kunanusont2017bandit}.

In this artificial model, $\x$ is treated as an $n$-bit binary number, and the true winning rate of $\x$ is defined as 
\begin{equation}
\P_{win}(\x) = \frac{Value(\x)}{(2^n-1)},\label{eq:p}
\end{equation}
where $Value(\x)$ denotes the numeric value of $\x$ located between $0$ and $2^n-1$.
Thus, the outcome of a game is either loss ($0$) or win ($1$) with probability $\P_{win}(\x)$ (Eq. \ref{eq:p}), as formalised in Eq. \ref{eqn:nwr}.
\begin{equation}
    G(\x) = \left\{
   \begin{array}{r l}
      1 & \text{ with probability } \P_{win}(\x), \\
      0 & \text{ otherwise.}\label{eqn:nwr}
   \end{array}
   \right .
\end{equation}


While resampling a game $r$ times, the average outcome, $\frac{1}{r}\sum_{i=1}^{r} G(\x)$, is the winning rate measured over $r$ repetitions of the game.
By the Law of Large Numbers (LLN), for an $r$ large enough, $\frac{1}{r}\sum_{i=1}^{r} G(\x)$ converges to $\P_{win}(\x) $.

\section{Experimental Settings}\label{sec:experiment}
Both the RMHC and the (1+1)-EA with mutation probability $\frac{1}{n}$ ($n$ is the solution dimension) have been applied for optimising each of the two problems using each of the two evaluation methods.
In each of the following two cases we measure the percentage of times that the optimiser returns the optimal solution.  
\begin{enumerate}[label=(\subscript{E}{\arabic*})]
\item \label{c1} We use the FHT such that the first time the optimiser visits the optimum, it returns that as its solution. If it uses the entire evaluation budget then it returns its current solution.
\item \label{c2} The optimiser returns its current solution when the evaluation budget has been used up, without prior knowledge of the optimal solution.
\end{enumerate}
Note that case \ref{c1} assumes that the optimal solution is already known.
During evaluation, each solution is evaluated $r$ times where $r$ varies from $1$ to $50$.  We call $r$ the resampling rate.  
The pseudocode of RMHC and (1+1)-EA with resamplings are given in Algorithms \ref{algo:rmhc} and \ref{algo:opo}. 
Note that in each algorithm the parent is re-evaluated at each iteration.
Each experiment has been repeated 10,000 times using uniformly randomly initialised solutions.  We chose the number of dimensions $n=10$.  This
small number of dimensions helps to emphasise the difference in the results of the alternative evaluation methods, though
differences are also present for higher-dimensional problems.  The evaluation budget (i.e. the number of raw fitness evaluations allowed) was set to $T=500$ for all experiments.
\begin{algorithm}[ptb]
\caption{\label{algo:rmhc}Random Mutation Hill Climber (RMHC) with resampling rate $r\in \N^+$ to maximise a noisy binary problem. $n$ is the problem dimension. $fitness^{(i)}(\x)$ refers to the $i^{th}$ call to the noisy fitness function for the given $\x$.}
\begin{algorithmic}
\State{Uniformly randomly initialise $x = (x^{1}, x^{2}, \dots, x^{n}) \in \{0,1\}^n$}
\While{Stopping condition not met}\Comment{\ref{c1} or \ref{c2}}
\State{$x'=x$}
\State{Uniformly randomly select dimension $d\in \{1,\dots,n\}$}
\State{$x'^{d} = 1-x'^{d}$} 
\If{$\frac1r \sum_{i=1}^{r} fitness^{(i)}(x') \geq \frac1r \sum_{i=1}^{r} fitness^{(i)}(x)$}
\State{$x=x'$}
\EndIf
\EndWhile
\State{\textbf{return} x}
\end{algorithmic}
\end{algorithm}
\begin{algorithm}[ptb]
\caption{\label{algo:opo}(1+1)-EA with mutation probability $\frac1n$ and resampling rate $r\in \N^+$ to maximise a noisy $n$-dimensional binary problem. $fitness^{(i)}(\x)$ refers to the $i^{th}$ call to the noisy fitness function for the given $\x$.}
\begin{algorithmic}
\State{Uniformly randomly initialise $x = (x^{1}, x^{2}, \dots, x^{n}) \in \{0,1\}^n$}
\While{Stopping condition not met}\Comment{\ref{c1} or \ref{c2}}
\State{$x'=x$} 
\For{$d \in \{1,\dots,n\}$}
\If{$random \leq \frac1n$} 
\State{$x'^{d} = 1-x'^{d}$} 
\EndIf
\EndFor
\If{$\frac1r \sum_{i=1}^{r} fitness^{(i)}(x') \geq \frac1r \sum_{i=1}^{r} fitness^{(i)}(x)$}
\State{$x=x'$}
\EndIf
\EndWhile
\State{\textbf{return} x}
\end{algorithmic}
\end{algorithm}

\section{Results\label{sec:res}}

Figures \ref{fig:omax} and \ref{fig:pmax} plot the percentage of times that an optimal solution is returned by the (1+1)-EA (left) and the RMHC (right). We focus on a case where the search space is small (solution dimension $n=10$), but the evaluation budget is also small. The details of the algorithm are identical except for the stopping conditions presented in \ref{c1} and \ref{c2} respectively (Section \ref{sec:experiment}). To highlight the difference between results using stopping conditions \ref{c1} and \ref{c2}, the figures of the best and worst success rates in each experiment are summarised in Table~\ref{tab}.
\begin{table}
\centering
\begin{tabular}{ccccc}
\hline
Algorithm & \multicolumn{2}{c}{(1+1)-EA} & \multicolumn{2}{c}{RMHC} \\
Stopping rule & \ref{c1} & \ref{c2} & \ref{c1} & \ref{c2}\\
\hline
Noisy OneMax & 85.61\% (1) & 38.98\% (5) & 97.44\% (1) & 65.44\% (7) \\
Noisy PMax & 19.56\% (1) & 1\% (16) & 28.67\% (1)  & 1.14\%(14)\\
\hline
\end{tabular}
\caption{\label{tab}The best success rates using different resampling rates over 10,000 optimisation trials by the (1+1)-EA and the RMHC on each of the two problems, using different stopping conditions.  The corresponding resampling rate is given in the parenthesis.}
\end{table}

\subsection{Noisy OneMax problem}
\begin{figure*}[htbp]
\centering
\begin{subfigure}[t]{.49\textwidth}
\includegraphics[width=1\textwidth]{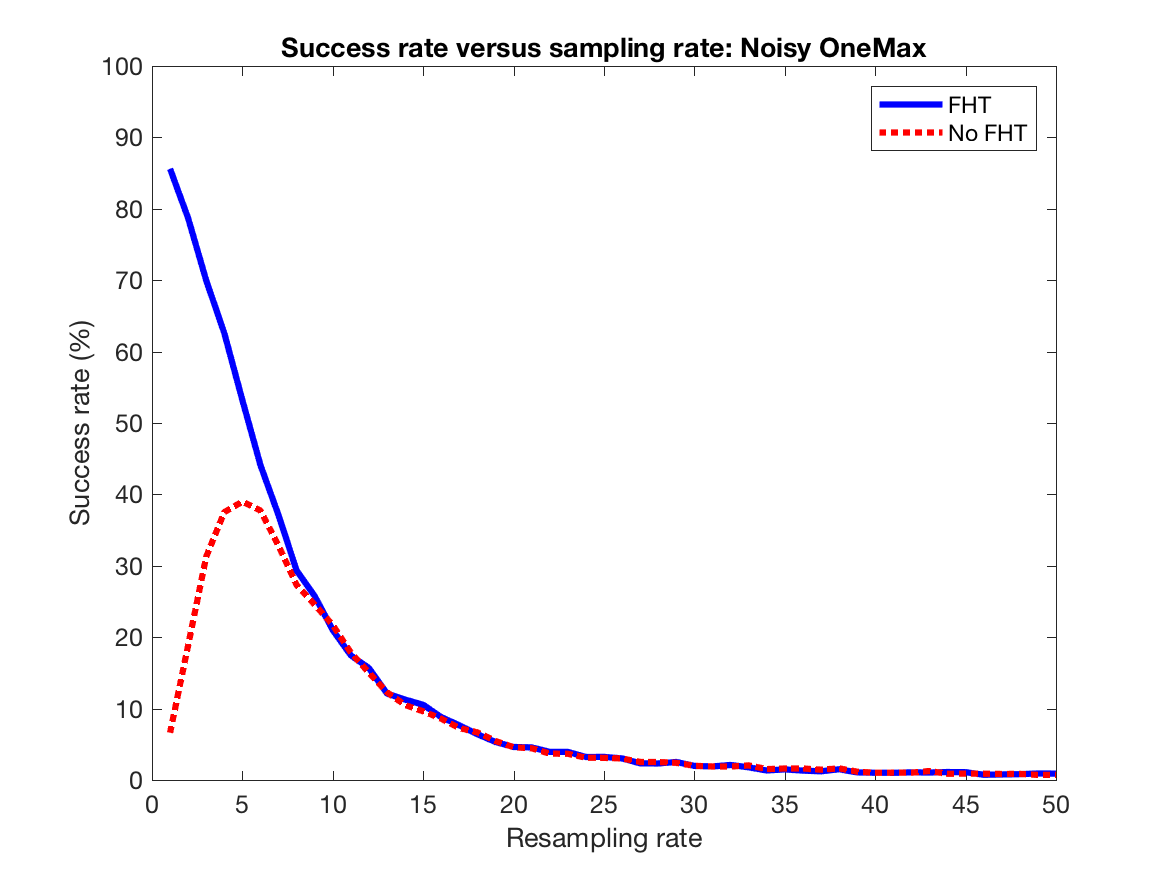}
\caption{\label{fig:omaxea}Optimised by (1+1)-EA with $p=0.1$.}
\end{subfigure}
\hfill
\begin{subfigure}[t]{.49\textwidth}
\includegraphics[width=1\textwidth]{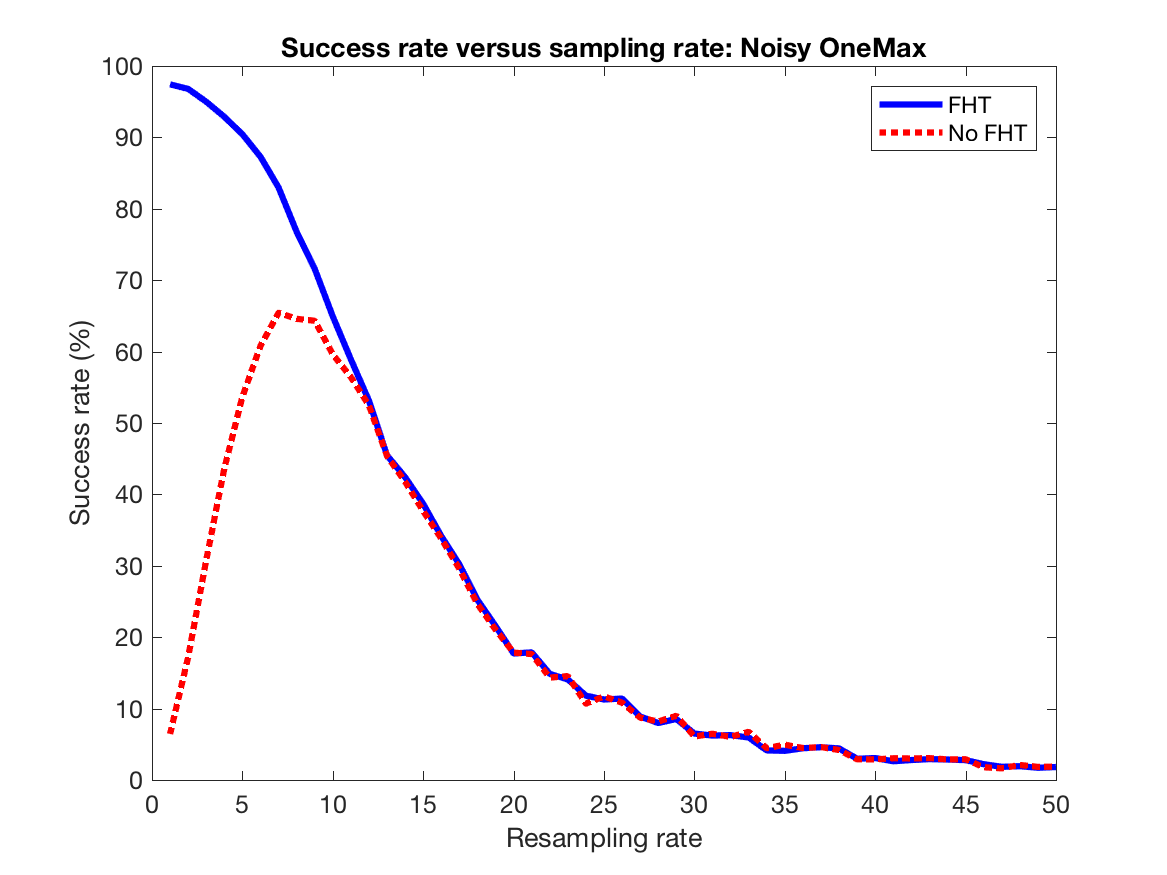}
\caption{\label{fig:omaxrmhc}Optimised by RMHC.}
\end{subfigure}
\caption{\label{fig:omax}Results on the 10-bit Noisy OneMax problem. 
\emph{x-axis:} resampling rate; \emph{y-axis:} percentage of times when the optimal solution is found over 10,000 trials.
The solid curves illustrate the success rate using the evaluation method \ref{c1}, while the dashed curves show the results using \ref{c2}. In both cases, each of the optimisation trials has started using a uniformly randomly initialised solution and a total budget of $T=500$ noisy fitness evaluations.}
\end{figure*}

When each algorithm operates for a fixed maximum number of fitness evaluations ($T=500$), the RMHC finds the optimal solution significantly more frequently than the (1+1)-EA (Figure \ref{fig:omax}).
Table \ref{tab} illustrates that the (1+1)-EA using \ref{c1} as the stopping condition gives the best success rate as $85.61\%$ using \ref{c1} but the true best success rate is only $38.98\%$ without assuming knowing the optimal solution in advance; the RMHC using \ref{c1} as the stopping condition gives the best success rate as $97.44\%$, dropping to $65.44\%$ if \ref{c2} is used.
Qian et al.~\cite{qian2014effectiveness} stated that resampling was useless in the 10-bit Noisy OneMax problem described in this paper and Liu et al.~\cite{liu2016optimal} concluded that in the same noise model, the optimal resampling rate for the RMHC increased with the problem dimension, and for the $10$-bit Noisy OneMax problem, the optimal resampling rate was 1. In both works~\cite{qian2014effectiveness,liu2016optimal}, the analysis was based on the condition that the optimisation terminated as soon as the optimum was found and there was no limit to the maximum number of fitness evaluations.
Indeed, in line with those predictions, Figure \ref{fig:omaxrmhc} shows that the optimal resampling rate for the RMHC, using the FHT and the maximum number of fitness evaluations as the stopping conditions, is 1, while the rate for the RMHC using only the maximum number of fitness evaluations as the stopping condition is 7.  The important point to note is that using FHT leads to a
false conclusion regarding both the optimal success rate and the corresponding optimal resampling rate.

\subsection{Noisy PMax problem}
\begin{figure*}[htbp]
\centering
\begin{subfigure}[t]{.49\textwidth}
\includegraphics[width=1\textwidth]{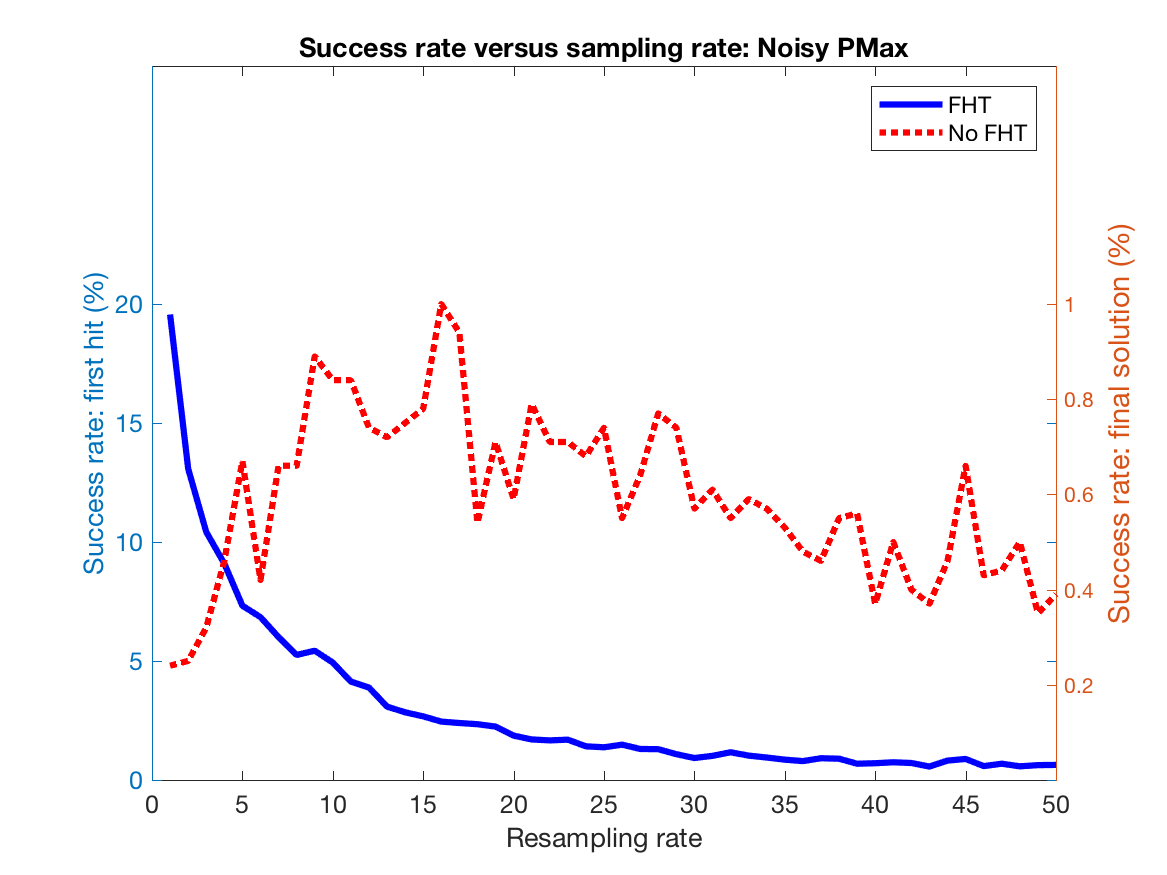}
\caption{\label{fig:perc}Optimised by (1+1)-EA with $p=0.1$.}
\end{subfigure}
\hfill
\begin{subfigure}[t]{.49\textwidth}
\includegraphics[width=1\textwidth]{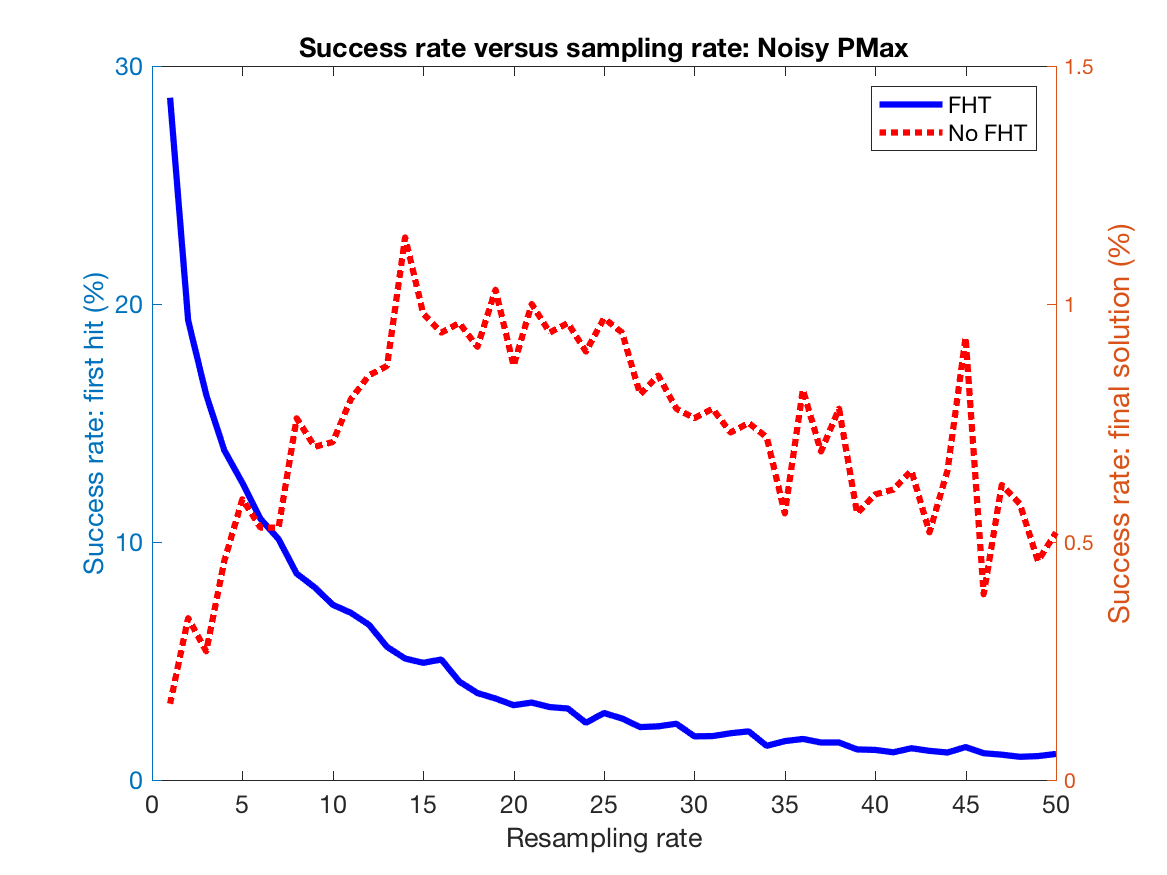}
\caption{\label{fig:fit}Optimised by RMHC.}
\end{subfigure}
\caption{\label{fig:pmax}Results on the 10-bit Noisy PMax problem (detailed in Section \ref{sec:gp}).  
\emph{x-axis:} resampling rate; 
\emph{left y-axis (solid curves):} percentage of times the optimal solution is found using the evaluation method \ref{c1} over 10,000 trials;
\emph{right y-axis (dashed curves):} percentage of times the optimal solution is found using the evaluation method \ref{c2} over 10,000 trials.
In both cases, each of the optimisation trials has started using a uniformly randomly initialised solution and a total budget of $T=500$ noisy fitness evaluations.}
\end{figure*}
Figure \ref{fig:pmax} shows the percentage of times that an optimal solution is returned over $10,000$ optimisation trials by the (1+1)-EA (left) and the RMHC (right) in the Noisy PMax problem described in Section \ref{sec:gp}, using the two stopping conditions discussed previously.
Table \ref{tab} illustrates that the (1+1)-EA using \ref{c1} as the stopping condition gives the best success rate as $19.56\%$, but the true best success rate (without the assumption that the optimal solution is known) is only $1\%$; the RMHC using \ref{c1} as the stopping condition gives the best success rate as $28.67\%$, but the true best success rate is only $1.14\%$.
The percentage of times that an optimal solution is returned using the stopping condition \ref{c1} is much higher than the percentage obtained when using the stopping condition \ref{c2}. However, having prior knowledge of the optimal solution and the access to the noise-free fitness is rare in real-world applications. Hence, the runtime analysis using the stopping condition \ref{c1} is too optimistic and unrealistic.

\section{Conclusion\label{sec:conc}}
To the best of the authors' knowledge, the current work around runtime analysis or prediction of the optimal resampling rate is based on the assumption that the optimal solution is known. Once the optimal noise-free fitness level is reached or the optimal solution is found, the optimisation stops immediately, thus the absorbing state is reached and the solution won't escape from the absorbing state.
In real-world applications, it is unlikely that we know the optimal solution in advance since there will be no reason to search for it.

The main conclusion is that First Hitting Time should not be used to evaluate the performance
of noisy optimisation algorithms since it can dramatically over-estimate the performance of
an algorithm, and may significantly under-estimate the amount of resampling required in
order to obtain optimal performance.  The conclusion is supported by the results of running
two simple but widely used optimisation algorithms on two different
test problems.

\ifCLASSOPTIONcaptionsoff
  \newpage
\fi



\balance
\bibliographystyle{IEEEtran}
\bibliography{main}
%

%




\end{document}